\def\year{2019}\relax
\documentclass[letterpaper]{article} 
\usepackage{aaai19}  
\usepackage{times}  
\usepackage{helvet}  
\usepackage{courier}  
\usepackage{url}  
\usepackage{graphicx}  
\usepackage{amsmath}
\usepackage{url}
\usepackage{multirow}
\usepackage{bm}
\usepackage{array}
\newcolumntype{L}[1]{>{\raggedright\let\newline\\\arraybackslash\hspace{0pt}}m{#1}}
\newcolumntype{C}[1]{>{\centering\let\newline\\\arraybackslash\hspace{0pt}}m{#1}}
\newcolumntype{R}[1]{>{\raggedleft\let\newline\\\arraybackslash\hspace{0pt}}m{#1}}

\DeclareMathOperator*{\argmin}{arg\,min}
\begin{document}
%
\title{A comprehensive empirical analysis on cross-domain semantic enrichment for detection of depressive language}
\author{
Nawshad Farruque, Randy Goebel and Osmar Za\"{i}ane\\
Department of Computing Science\\
University of Alberta\\
Alberta, T6G 2E8, Canada\\
}
\maketitle
\begin{abstract}

We analyze the process of creating word embedding feature representations designed for a learning task when annotated data is scarce, for example, in depressive language detection from Tweets. We start with a rich word embedding pre-trained from a large general dataset, which is then augmented with embeddings learned from a much smaller and more specific domain dataset through a simple non-linear mapping mechanism. We also experimented with several other more sophisticated methods of such mapping including, several auto-encoder based and custom loss-function based methods that learn embedding representations through gradually learning to be close to the words of similar semantics and distant to dissimilar semantics.  Our strengthened representations better capture the semantics of the depression domain, as it combines the semantics learned from the specific domain coupled with word coverage from the general language. We also present a comparative performance analyses of our word embedding representations with a simple bag-of-words model, well known sentiment and psycholinguistic lexicons, and a general pre-trained word embedding.  When used as feature representations for several different machine learning methods, including deep learning models in a depressive Tweets identification task, we show that our augmented word embedding representations achieve a significantly better F1 score than the others, specially when applied to a high quality dataset. Also, we present several data ablation tests which confirm the efficacy of our augmentation techniques.
\end{abstract}
\section{Introduction}
Depression or Major Depressive Disorder (MDD) is regarded as one of the most commonly identified mental health problems among young adults in developed countries, accounting for 75\% of all psychiatric admissions \cite{boyd1982screening}. Most people who suffer from depression do not acknowledge it, for various reasons, ranging from social stigma to plain ignorance; this means that a vast majority of depressed people remain undiagnosed.
Lack of diagnosis eventually results in suicide, drug abuse, crime and many 
other societal problems. For example, depression has been found to be a major cause behind 
800,000 deaths committed through suicide each year 
worldwide\footnote{\url{https://who.int/mental_health/prevention/suicide/suicideprevent/en/}}. Moreover, the economic burden created by depression is estimated to 
have been 210 billion USD in 2010 in the USA alone \cite{greenberg2015economic}. Hence, detecting, monitoring and treating depression is very important and there is huge need for effective, inexpensive and almost real-time 
interventions. In such a scenario, social media, such as, Twitter and Facebook, provide the foundation of a remedy. Social media are very popular among young adults, where depression is prevalent. In addition, it has been found that people who are otherwise socially aloof (and more prone to having depression) can be very active in the social media platforms \cite{Choudhury2013Pred}. 
As a consequence, there has been significant depression detection research, based on various social media attributes, such as social network size, social media behavior, and language used in social media posts. Among these multi-modalities, human language alone can be a very good predictor of depression \cite{Choudhury2013Pred}. However, the main bottle neck of social media posts-based depression detection task is the lack of labelled data to identify rich feature representations, which provide the basis for constructing models that help identify depressive language.

Here we discuss the creation of a word embedding  
that leverages both the Twitter vocabulary (from pre-trained Twitter word embedding) and depression semantics (from a word embedding created from depression forum posts) to identify depressive Tweets. We believe our proposed methods would significantly relieve us from the burden of curating a huge volume of human annotated data / high quality labelled data (which is very expensive and time consuming) to support the learning of better feature representations, and eventually lead to improved classification. In the next sections we provide a brief summary of earlier research together with some background supporting our formulation of our proposed methods for identifying depression from Tweets. 

Throughout our paper, we use the phrase ``word embedding'' as an object that consists of word vectors. So by ``word embeddings'' we mean multiple instances of that object.

\section{Background \& motivation}
Previous studies suggest that the words we use in our daily life can express our mental state, mood and emotion \cite{Pennebaker2003}. Therefore analyzing language to identify and monitor human mental health problems has been regarded as an appropriate avenue of mental health modeling. With the advent of social media platforms, researchers have found that social media posts can be used as a good proxy for our day to day language usage \cite{Choudhury2013Pred}. There have been many studies that identify and monitor depression through social media posts in various social media, such as, Twitter \cite{reece2017forecasting,Choudhury2013RoleSM,Choudhury2013Pred}, Facebook \cite{schwartz2014towards,moreno2011feeling} and online forums \cite{yates2017depression}. 

	
Depression detection from social media posts can be specified as a low resource supervised classification task 
because of the paucity of valid data. Although there is no concrete precise definition of valid data, previous research emphasizes collecting social media posts, which are either validated by annotators as carrying clues of depression, or coming from the people who are clinically diagnosed as depressed, or both. Based on the methods of depression intervention using these data, earlier research can be mostly divided into two categories: (1) general categories of post-specific depression detection (or depressive language detection) \cite{Choudhury2013RoleSM,jamil2017monitoring,vioules2018detection}, and (2) user-specific depression detection, which considers all the posts made by a depressed user in a specific time window \cite{Resnik2013,resnik2015beyond}. The goal of (1) is to identify depression in a more fine grained level, i.e., in social media posts, which further helps in identifying depression inclination of individuals when analyzed by method (2). 
    

For the post specific depression detection task, previous research concentrate on the extraction of depression specific features used to train machine learning models, e.g., building depression lexicons based on unigrams present in posts from depressed individuals \cite{Choudhury2013Pred}, depression symptom related unigrams curated from depression questionnaires \cite{cheng2016psychologist},  metaphors used in depressive language \cite{neuman2012proactive}, or psycholinguistic features in LIWC \cite{tausczik2010psychological}.  For user specific depression identification, variations of topic modeling have been popular to identify depressive topics and use them as features \cite{resnik2015beyond,Resnik2013}. But recently, some research has used convolutional neural network (CNN) based deep learning models to learn feature representations \cite{yates2017depression} and \cite{orabi2018deep}. Most deep learning approaches require a significant volume of labelled data to learn the depression specific embedding from  scratch, or from a pre-trained word embedding in a supervised manner. So, in general, both post level and user level depression identification research emphasize the curation of labelled social media posts indicative of depression, which is a very expensive process in terms of time, human effort, and cost. Moreover, previous research showed that a robust post level depression 
identification system is an important prerequisite for accurately identifying depression at the user level \cite{Choudhury2013RoleSM}. In addition, most of this earlier research leveraged Twitter posts to identify depression because a huge volume of Twitter posts are publicly available. 

Therefore the motivation of our research comes from the need for a better feature representation specific to depressive language, and reduced dependency on a large set of (human annotated) labelled data for depressive Tweet detection task. We proceed as follows:

\begin{enumerate}
  \item We create a word embedding space that encodes the semantics of depressive language from a 
  small but high quality 
  depression corpus curated from depression related public forums.
  \item We use that word embedding to create feature representations for our Tweets and feed them to our machine learning models to identify depressive Tweets; this achieves good accuracy, even with very small amount of labelled Tweets.
  \item Furthermore, 
we adjust a pre-trained Twitter word embedding based on our depression specific word embedding, using a non-linear mapping between the embeddings (motivated by the work of \cite{mikolov2013exploiting} and \cite{smith2017offline} on bilingual dictionary induction for machine translation), and use it to create feature representation for our Tweets and feed them to our machine learning models. This helps us achieve 4\% higher F1-score than our strongest baseline in depressive Tweets detection.
\end{enumerate}

Accuracy improvements mentioned in points 2 and 3 above are true for a high quality dataset curated through rigorous human annotation, 
as opposed to the low quality dataset with less rigorous human annotation; this indicates the effectiveness of our proposed feature representations for depressive Tweets detection. To the best of our knowledge, ours is the first effort to build a depression specific word embedding for identifying depressive Tweets, and to formulate a method to gain further improvements on top of it, then to present a comprehensive analysis on the quantitative and qualitative performance of our embeddings.
    
\section{Datasets} \label{sec:datasets}
Here we provide the details of our two datasets that we use for our experiments and their annotation procedure, the corpus they are curated from and their quality comparisons. 

\subsection{Dataset1} Dataset1 is curated by the ADVanced ANalytics for data SciencE (ADVANSE) research team at the University of Montpellier, France \cite{vioules2018detection}. This dataset contains Tweets having key-phrases generated from the American Psychiatric Association (APA)'s list of risk factors and the American Association of Suicidology (AAS)'s list of warning signs related to suicide. Furthermore, they randomly investigated the authors of these Tweets to identify 60 distressed users who frequently write about depression, suicide and self mutilation. They also randomly collected 60 control users.
Finally, they curated a balanced and human annotated dataset of a total of around 500 Tweets, of which 50\% Tweets are from distressed and 50\% are from control users, with the help of seven annotators and one professional psychologist. The goal of their annotation was to provide a distress score (0 - 3) for each Tweet. They reported a Cohen's kappa agreement score of 69.1\% for their annotation task. Finally, they merged Tweets showing distress level 0, 1 as control Tweets and 2, 3 as distressed Tweets. \textit{Distressed Tweets} carry signs of suicidal ideation, self-harm and depression while control Tweets are about daily life occurrences, such as weekend plans, trips and common distress such as exams, deadlines, etc. We believe this dataset is perfectly suited for our task, and we use their distressed Tweets as our depressive Tweets and their control as our control. 

\subsection{Dataset2}
Dataset2 is collected by a research group at the University of Ottawa \cite{jamil2017monitoring}.  They first filtered depressive Tweets from \#BellLetsTalk2015 (a Twitter campaign) based on keywords such as, suffer, attempt, suicide, battle, struggle and first person pronouns. 
Using topic modeling, they removed Tweets under the topics of public campaign, mental health awareness, and raising money. They further removed Tweets which contain mostly URLs and are very short. Finally, from these Tweets they identified 30 users who self-disclosed their own depression, and 30 control users who did not. They 
employed two annotators to label Tweets from 10 users as either depressed or non-depressed. They found that their annotators labelled most Tweets as non-depressed. To reduce the number of non-depressive Tweets, they further removed neutral Tweets from their dataset, as they believe neutral Tweets surely do not carry any signs of depression. After that, they annotated Tweets from the remaining 50 users with the help of two annotators with a Cohen's kappa agreement score of 67\%. Finally, they labelled a Tweet as depressive if any one of their two annotators agree, to gather more depressive Tweets. This left them with 8,753 Tweets with 706 depressive Tweets.

\subsection{Quality of Datasets} Here we present a comparative analysis of our datasets based on their curation process and the linguistic components present in them relevant to depressive language detection as follows:
\subsubsection{Analysis based on data curation process:}
We think Dataset2 is of lower quality compared to Dataset1 for the following reasons: (1) this dataset is collected from the pool of Tweets which is a part of a mental health campaign, and thus compromises the authenticity of the Tweets; (2) the words they used for searching depressive Tweets are not validated by any depression or suicide lexicons; (3) although they used two annotators (none of them are domain experts) to label the Tweets, they finally considered a Tweet as depressive if at least one annotator labelled it as depressive, hence introduced more noise in the data;  (4) it is not confirmed how they identified neutral Tweets since their neutral Tweets may convey depression as well; (5) they identified a person is depressed if s/he disclose their depression, but they did not mention how they determined these disclosures. Simple regular expression based methods to identify these self disclosures can introduce a lot of noise in the data. In addition, these self disclosures may not be true.

\subsubsection{Analysis based on linguistic components present in the dataset:} \label{subsect: LIWC}
\begin{figure}[!ht]
\centering
\includegraphics[width=0.3 \textwidth]{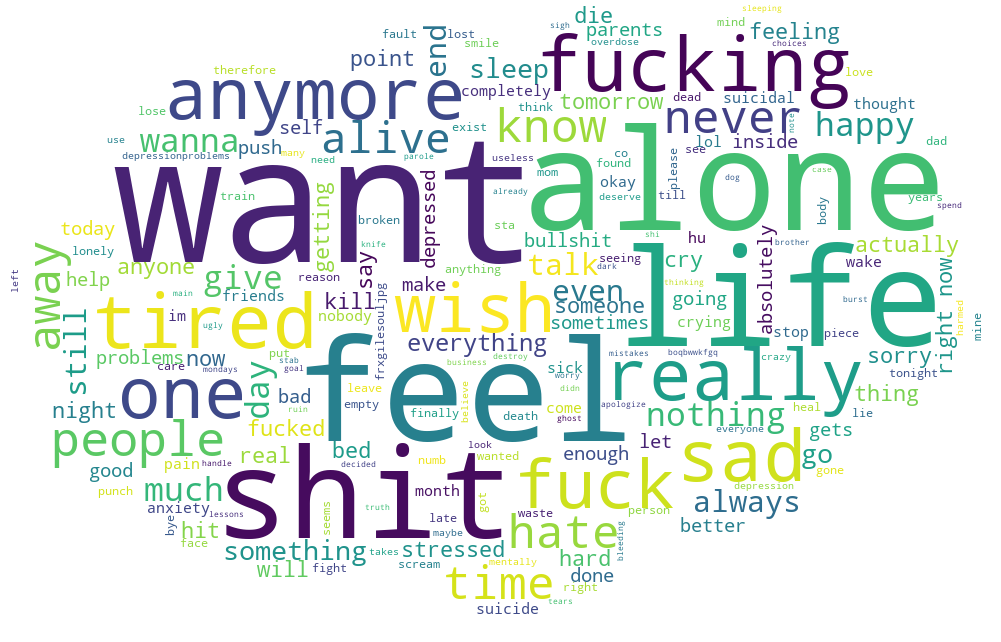}
\caption{\label{fig:wordcloud1} Dataset1 depressive Tweets word cloud }
\end{figure}

\begin{figure}[!ht]
\centering
\includegraphics[width=0.3 \textwidth]{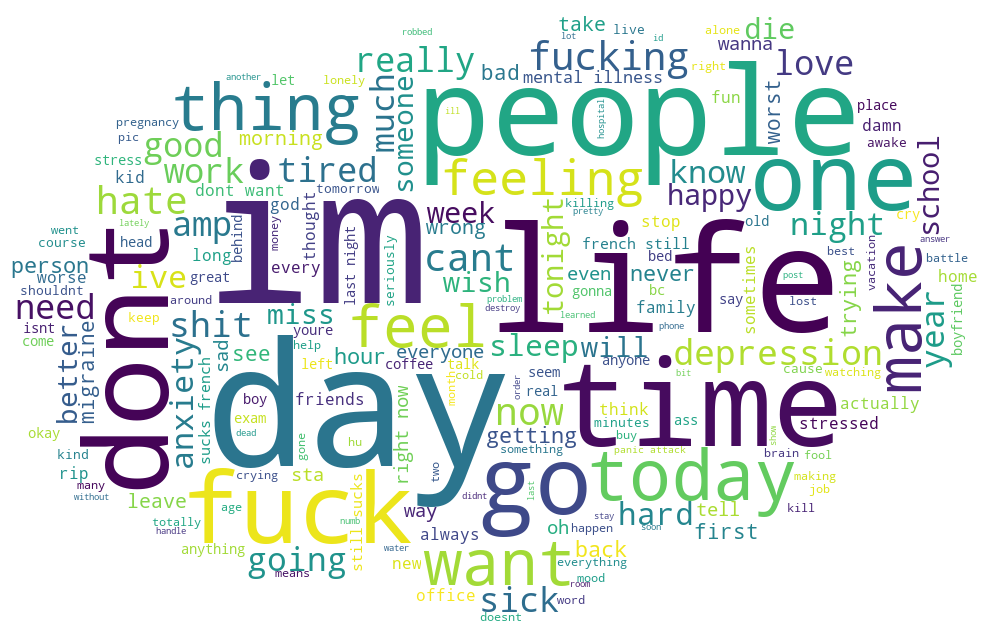}
\caption{\label{fig:wordcloud2} Dataset2 depressive Tweets word cloud}
\end{figure}

For this analysis, we use Linguistic Inquiry and Word Count (LIWC) \cite{tausczik2010psychological}. LIWC is a tool widely used in psycholinguistic analysis of language. It extracts the percentage of words in a text, across 93 pre-defined categories, e.g., affect, social process, cognitive processes, etc. To analyse the quality of our datasets, we provide scores of few dimensions of LIWC lexicon relevant for depressive language detection \cite{nguyen2014affective}, \cite{Choudhury2013Pred} and \cite{kuppens2012emotional}, such as, 1st person pronouns, anger, sadness, negative emotions, etc, see Table \ref{tab:LIWC-feats} for the depressive Tweets present both in our datasets. The bold items in that table shows significant score differences in those dimensions for both datasets and endorses the fact that Dataset1 indeed carries more linguistic clues of depression than Dataset2 (the higher the score, the more is the percentage of words from that dimension is present in the text). Moreover, Tweets labelled as depressive in Dataset2 are mostly about common distress of everyday life unlike those of Dataset1, which are indicative of severe depression. Figures \ref{fig:wordcloud1} and \ref{fig:wordcloud2} depict the word clouds created from Dataset1 and Dataset2 depressive Tweets respectively. We provide few random samples of Tweets from Dataset1 and Dataset2 at Table \ref{tab:exampleTweets} as well. 



\begin{table}[!ht]
\small
\centering
\begin{tabular}{|C{1.5cm}|C{1.7cm}|C{1.9cm}|C{1.9cm}|}
\hline
\textbf{LIWC Category} & \textbf{Example Words} & \textbf{Dataset1 Depressive Tweets score}  & \textbf{Dataset2 Depressive Tweets score} \\ \hline
\textbf{1st person pronouns} &I, me, mine &12.74 &7.06\\ \hline
\textbf{Negations} &no, not, never &3.94 &2.63 \\ \hline
Positive Emotion &love, nice, sweet &2.79 &2.65 \\ \hline
\textbf{Negative Emotion} &hurt, ugly, nasty &8.59 &6.99 \\ \hline
Anxiety &worried, fearful &0.72 &1.05 \\ \hline
Anger &hate, kill, annoyed &2.86 &2.51 \\ \hline
\textbf{Sadness} &crying, grief, sad &3.29 &1.97 \\ \hline
Past Focus &ago, did, talked &2.65 &3 \\ \hline
\textbf{Death} &suicide, die, overdosed &1.43 &0.44 \\ \hline
\textbf{Swear} &fuck, damn, shit &1.97 &1.39 \\ \hline
\end{tabular}
\caption{Score of Dataset1 and Dataset2 in few LIWC dimensions relevant to depressive language detection}
\label{tab:LIWC-feats}
\end{table}

\begin{table}[!ht]
\small
\centering
\begin{tabular}{ |l|p{6cm}| }
\multicolumn{1}{ c }{}\\
\hline
\textbf{Datasets} & \textbf{Depressive Tweets} \\ \hline
\multirow{1}{*}{Dataset1} & ``I wish I could be normal and be happy and feel things like other people'' \\ \cline{2-2}
 &  ``I feel alone even when I'm not'' \\ \cline{2-2}
 &  ``Yesterday was difficult...and so is today and tomorrow and the days after...'' \\
 \hline
\multirow{1}{*}{Dataset2} & ``Last night was not a good night for sleep... so tired And I have a gig tonight... yawnnn'' \\ \cline{2-2}
 & ``So tired of my @NetflixCA app not working, I hate Android 5''  \\ \cline{2-2}
 & ``I have been so bad at reading Twitter lately, I don't know how people keep up, maybe today I'll do better'' \\
 \hline
\end{tabular}
\caption{Sample Tweets from Dataset1 and Dataset2}
\label{tab:exampleTweets}
\end{table}

\subsection{Why these two datasets?}
We believe these two datasets represent the two broad categories of publicly available Twitter datasets for the depressive Tweets identification task. One category relies on keywords related to depression and suicidal ideation to identify depressive Tweets and employed rigorous annotation to filter out noisy Tweets (like Dataset1); the other relies on self disclosures of Twitter users to identify depressive Tweets and employed less rigorous annotation (like Dataset2) to filter noisy Tweets. So any other datasets that fall into one of above categories or do not go through annotation procedure atleast like Dataset2, such as datasets released in a CLPsych 2015 shared task \cite{coppersmith2015clpsych} are not evaluated in this research. Moreover, our Dataset2 is a representative of imbalanced dataset (with fewer depressive Tweets than non-depressive Tweets) which is a very common characteristic of the datasets for depressive Tweets identification task. It is also to be noted that we are interested in depressive Tweets identification, so our datasets are from Twitter, not from depression forums. We are using depression forum posts only to learn improved word embedding feature representation that can help us identifying depressive Tweets in the above mentioned Twitter datasets.

\subsection{Creating a depression specific corpus}
\label{subsec: depcorpora}
To build a depression specific word embedding, we curate our own depression corpus.  
For this, we collect all the posts from the Reddit depression forum: r/depression \footnote{\url{reddit.com/r/depression/}} between 2006 to 2017 and all those from Suicidal Forum \footnote{\url{suicideforum.com/}} and concatenated for a total of 856,897 posts. We choose these forums because people who post anonymously in these forums usually suffer from severe depression and share their struggle with depression and its impact in their personal lives \cite{de2014mental}. We believe these forums contain useful semantic components indicative of depressive language. 


\section{Feature extraction methods} \label{sec:features}
\subsection{Bag-of-Words (BOW)} \label{subsec:bow} We represent each Tweet as a vector of vocabulary terms and their frequency counts in that Tweet, also known as bag-of-words. The vocabulary terms refer to the most frequent 400 terms existing in the training set. Before creating the vocabulary and the vector representation of the Tweets, we perform the following preprocessing: (1) we make the Tweets all lowercase, then (2) tokenize them using the NLTK Tweet tokenizer \footnote{\url{nltk.org/api/nltk.tokenize.html}}; the reason for using Tweet tokenizer is to consider Tweet emoticons (:-)), hashtags (\#Depression) and mentions (@user) as single tokens; we then (3) remove all stop words except the first person pronouns such as, I, me and my (because they are useful for depression detection) and then (4) use NLTK porter stemmer \footnote{\url{nltk.org/_modules/nltk/stem/porter.html}}. 
Stemming helps us reduce sparsity of the bag-of-words representations of Tweets. 

\subsection{Lexicons} We experimented with several emotion and sentiment lexicons, such as, LabMT \cite{dodds2011temporal}, Emolex \cite{mohammad2013nrc}, AFINN \cite{nielsen2011new}, LIWC \cite{tausczik2010psychological}, VADER \cite{gilbert2014vader}, NRC-Hashtag-Sentiment-Lexicon (NHSL) \cite{kiritchenko2014sentiment},  NRC-Hashtag-Emotion-Lexicon (NHEL) \cite{COIN:COIN12024} and CBET \cite{shahrakilexical}. Among these lexicons we find LIWC and NHEL perform the best and hence we report the results of these two lexicons. The following subsections provide a brief description of LIWC, NHEL and lexicon-based representation of Tweets.

\subsubsection{Linguistic Inquiry and Word Count (LIWC):} 
LIWC has been widely used as a good baseline for depressive Tweet detection in earlier research \cite{nguyen2014affective,coppersmith2014quantifying}. We use it to convert a Tweet into a fixed length vector representation of 93 dimensions, that is then used as the input for our machine learning models. 

\subsubsection{NRC Hashtag Emotion Lexicon (NHEL):} 
In NHEL there are 16,862 unigrams, each of which are associated with a vector of 8 scores for 8 emotions, such as, anger, anticipation, disgust, fear, joy, sadness, surprise and trust. Each of the scores (a real value between $0$ and $\infty$) indicate how much a particular unigram is associated with each of the 8 emotions.
In our experiments, we tokenize each Tweet as described in the Bag-of-Words (BOW) section, then we use the lexicon to determine a score for each token in the Tweet; finally, we sum them to get a vector of 8 values for each Tweet, which represents the expressed emotions in that tweet and their magnitude. Finally, we use that value as a feature for our machine learning models.

\subsection{Distributed representation of words} \label{subsec:wordembed} 
Distributed representation of words (also known as word embedding (WE) or a collection of word vectors \cite{Mikolov2013}) capture the semantic and syntactic similarity between a word and its context defined by its neighbouring words that appear in a fixed window, and has been successfully used as a compact feature representation in many downstream NLP tasks.
Previous research show that a domain specific word embedding is usually better for performing domain specific tasks than a general word embedding, e.g., \cite{bengio2014word} proposed word embedding for speech recognition, \cite{tang2014learning} proposed the same for sentiment classification and \cite{asgari2015continuous} for representing biological sequences. Inspired by these works, we here report the construction of depression specific word embedding, 
in an unsupervised manner, 
In addition, we report that the 
word embedding resulting from a non-linear mapping between general (pre-trained) word embedding and depression specific word embedding can be a very useful feature representation for our depressive Tweet identification task. 

Our embedding adjustment method has some similarity to embedding retrofitting proposed by \cite{faruqui:2014:NIPS-DLRLW} and embedding refinement proposed by \cite{yu2018refining}, in the sense that we also adjusted (or retrofitted) our pre-trained embedding. However, there is a major difference between their method and ours. They only adjusted those words, $w \in V$, where $V$ is the common vocabulary between their pre-trained embedding and semantic/sentiment lexicons, e.g. WordNet \cite{miller1995wordnet}, ANEW \cite{nielsen2011new} etc. By this adjustment they brought each word in the pre-trained embedding closer to the other words which are semantically related to them (as defined in the semantic lexicons) through an iterative update method, where all these words are member of $V$. So their method strictly depends on semantic lexicons and their vocabularies. In depression detection research, where labelled data is scarce and human annotation is expensive, building depression lexicons (given there is no good publicly available depression lexicons) and using them for retrofitting is counter intuitive. Even if we create one, there is a good chance that its vocabulary would be limited.  Also, most importantly, there is no comprehensive discussion/analysis in their paper on how to retrofit those words which are only present in pre-trained embedding but not in semantic lexicons or out-of-vocabulary (OOV) words. 
In our method we do not depend on any such semantic lexicons. We retrofitted a general pre-trained embedding based on the semantics present in depression specific embedding through a non-linear mapping between them. Our depression specific embedding is created in an unsupervised manner from depression forum posts. Moreover, through our mapping process we learn a transformation matrix, see Equation \ref{eq:3}, that can be further used to predict embedding for OOVs and this helps us to achieve better accuracy (see Table \ref{tab:embed-augment}).


Interestingly, there have been no attempts taken in depressive language detection research area which primarily focus on building better depression specific embedding in an unsupervised manner, then further analyse its use in augmenting a general pre-trained embedding. Very recently \cite{orabi2018deep} proposed a multi-task learning method which learns an embedding in a purely supervised way by simultaneously performing (1) adjacent word prediction task from one of their labelled train dataset and (2) depression/PTSD sentence prediction task again from the same labelled train dataset, where this labelled dataset was created with the help of human annotation. We have no such dependency on labelled data. Also, we have disjoint sets of data for learning/adjusting our embedding and detecting depressive Tweets across all our experiments, unlike them, which makes our experiments fairer than theirs. They did not provide any experimental result and analysis on depressive post (or Tweet) identification rather on depressive Twitter user identification which is fundamentally different from our task. Also, our paper discusses the transferability of depressive language specific semantics from forums to microblogs, which is not the focus of their paper. 
Finally, we argue that the dataset they used for depression detection is very noisy and thus not very suitable for the same (See ``Quality of Datasets'' section). 

In the following subsections we describe different word embeddings used in our experiments. 

\subsubsection{General Twitter word Embedding (TE):} We use a pre-trained 400 dimensional skip-gram word embedding learned from $400$ million Tweets with vocabulary size of  $3,039,345$ words \cite{godin2015multimedia} as a representative of word embedding learned from a general dataset (in our case, Tweets); we believe this captures the most relevant vocabulary for our task. The creator of this word embedding used negative sampling ($k = 5$) with a context window size = $1$ and mincount = $5$.  Since it is pre-trained, we do not have control over the parameters it uses and simply use it as is. 

\subsubsection{Depression specific word Embedding (DE):} We create a 400 dimensional depression specific word embedding (DE) on our curated depression corpus. 
First, we identify sentence boundaries in our corpora based on punctuation, such as: ``?'',``!'' and ``.''. 
We then feed each sentence into 
a skip-gram based word2vec implementation in gensim
\footnote{\url{radimrehurek.com/gensim/models/word2vec.html}}. We use negative sampling ($k=5$) with the context window size = $5$ and mincount = $10$ for the training of these word embeddings. DE has a vocabulary size of $29,930$ words. We choose skip-gram for this training because skip-gram learns good embedding from a small corpus \cite{mikolov2013efficient}.


\subsubsection{Adjusted Twitter word Embedding (ATE): a non-linear mapping between TE and DE:} 
\begin{figure}[!ht]
\centering
\includegraphics[width=0.5 \textwidth] {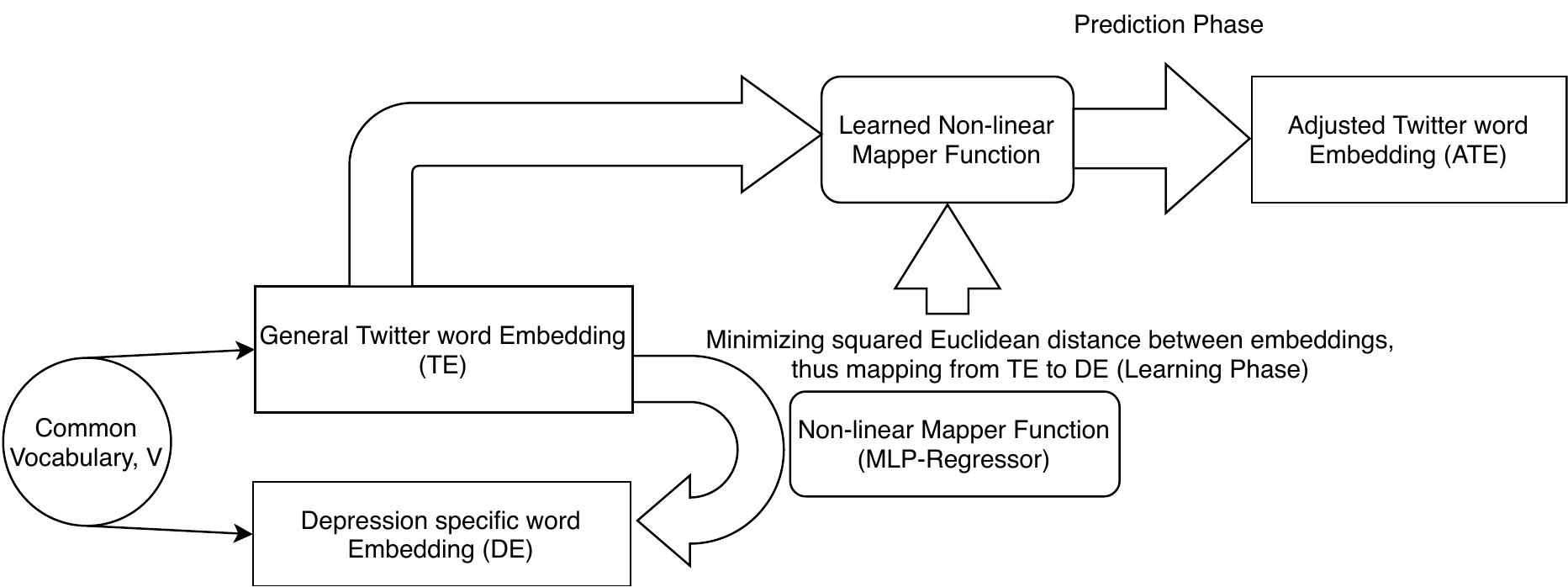} 
\caption{\label{fig:Mapper} Non-linear mapping of TE to DE (creation of ATE)}
\end{figure}

In this step, we create a non-linear mapping between TE and DE. 
To do this, we use a Multilayer Perceptron Regressor (MLP-Regressor) with a single hidden layer with 400 hidden units and Rectified Linear Unit (ReLU) activations (from hidden to output layer), which attempts to minimize the Minimum Squared Error (MSE) loss function, $\mathcal{F(\theta)}$ in Equation \ref{eq:1}, using stochastic gradient descent: 

\begin{equation}
\label{eq:1}
\mathcal{F(\theta)} = \argmin_\theta (\mathcal{L(\theta)})
\end{equation}
where
\begin{equation}
\label{eq:2}
\mathcal{L(\theta)} = \frac{1}{m}\sum_{i=1}^{m}||g_i(x) - y_i||_{2}^{2}
\end{equation} 
and
\begin{equation}
\label{eq:3}
g(x) = ReLU(b_1 + W_1(b_2+W_2x))
\end{equation} 
here, $g(x)$ is the non-linear mapping function between the vector $x$ (from TE) and $y$ (from DE) of a word $w \in V$, where, $V$ is a common vocabulary between TE and DE; $W_1$ and $W_2$ are the hidden-to-output and input-to-hidden layer weight matrices respectively, $b_1$ is the output layer bias vector and $b_2$ is the hidden layer bias vector (all these weights and biases are indicated as $\theta$ in Equation \ref{eq:1}) 
In Equation \ref{eq:2}, $m$ is the
length of $V$ (in our case it is 28,977). 
Once the MLPR learns the $\theta$ that minimizes $\mathcal{F(\theta)}$, 
it is used to predict the vectors for the words in TE which are not present in DE (i.e., out of vocabulary(OOV) words for DE). After this step, we finally get an adjusted Twitter word embedding 
which encodes the semantics of depression forums as well as word coverage from Tweets. We call these embedding the Adjusted Twitter word Embedding (ATE). The whole process is depicted in Figure \ref{fig:Mapper}. 


\subsubsection{ATE original (ATE(orig.)):} We report our general Twitter embedding (TE) adjusted by DE(original) or DE(orig.) and we name it adjusted Twitter embedding, ATE(original) or ATE(orig.). DE(orig.) is created with same parameter settings as our TE. We show that our DE with more frequent words (trained with mincount=10) and a bit larger context (context window size=5) help us create an improved ATE. 


A summary of the vocabulary sizes and the corpus our embedding sets are built on is provided in Table \ref{tab:corpusStat}.

\begin{table}[!ht]
\small
\centering
\begin{tabular}{|C{2cm}|C{2cm}|c|c|}
\hline
\textbf{Word Embeddings} & \textbf{Corpus Type} & \textbf{\#Posts}  & \textbf{Vocab. Size} \\ \hline
TE, ATE and ATE (orig.) &Twitter &400M &3M\\ \hline
DE &Depression Forums &1.5M &30K \\ \hline
\end{tabular}
\caption{Corpus and vocabulary statistics for word embeddings}
\label{tab:corpusStat}
\end{table}

\subsubsection{Conditions for embedding mapping/adjustment:} Our non-linear mapping between two embeddings works better given that those two embeddings are created from the same word embedding creation algorithm (in our case skip-gram) and have same number of dimensions (i.e. 400). We also find that a non-linear mapping between our TE and DE produces slightly better ATE than a linear mapping for our task, although the former is a bit slower.

\subsection{Other embedding augmentation methods:} We experiment with two more embedding augmentation methods. These methods are actually a slightly complex extensions of our proposed methods and do not necessarily surpass them in accuracy, hence we did not report them in our `Results Analysis' section, rather discuss here briefly.

\subsubsection{ATE-Centroids:} 
We propose two more methods of embedding augmentation which can be seen as an extension to our ATE construction method. In general, we call these methods ATE-Centroids. For a word $w \in V$, where $V$ is the common vocabulary between TE and DE. We learn a non-linear mapper (like the one we use to create ATE) which does the following, (1) it learns to minimize the squared euclidean distance between the embedding of $w$ in TE and the centroid (or average of word vectors) calculated from $w$ and its neighbours in DE (i.e. the words which are close to it in Euclidean distance in DE), we name the resulting embedding from this method, ATE-Centroid-1. (2) along with (1), it learns to maximize the distance between $w$ in TE and the centroid calculated from the word vectors of the distant words from $w$ in DE, we name the resulting embedding from this method, ATE-Centroid-2. After learning, the mapper is used to predict OOV words (the words which are in TE but not in DE). So in summary, by doing operations (1) and (2), we basically adjust a source pre-trained embedding (in our case TE) by pushing its constituent words close to the words of close semantics and away from the words of distant semantics according to the semantics defined in our target embedding (in our case DE). 

These methods obtain average F1 scores which is 0.7\% below than our best models in our two datasets. The reason of this slight under-performance could be the fact that the mapping process is less accurate than our best models. The stability of F1 scores (i.e. the standard deviation) is not significantly different than our best models and around on average $0.020$ across our datasets. See Table \ref{tab:meta-embeddings} and \ref{tab:meta-embeddings-centroids}.

\subsubsection{Averaged Autoencoded Meta-embedding (AAEME):} We try a state-of-the-art variational autoencoder based meta-embedding \cite{bollegala2018learning} created from our TE and DE. In this method, an encoder encodes the word vector of, $w \in V$, where $V$ is the common vocabulary between TE and DE. Then the average of those encoded vectors is calculated, which is called ``meta-embedding'' of $w$ according to \cite{bollegala2018learning}. Finally, a decoder is used to re-construct the corresponding TE and DE vector of $w$ from that meta-embedding. This way we gradually learn the meta-embedding that is supposed to hold the useful semantics from TE and DE for all the words in $V$.

With this meta-embedding (which we call ATE-AAEME), we achieve F1 scores on average 3.45\% less than our best model in both datasets. Since this method works better with a bigger common vocabulary between embeddings, we learn meta-embedding from TE and ATE instead of DE, we call it ATE-AAEME-OOV. This slightly improves F1 score by 2\%, but still in both datasets the F1 score we achieve is 1.36\% (on average) less than our best models. Moreover, ATE-AAEME-OOV achieves 1.3\% less stable F1-scores than our best model in Dataset1 but only 0.09\% more stable F1 scores than our best model in Dataset2. So we observe that the performance of AAEME method is significantly dependent on an efficient mapper function that we outlined in this paper. 
See Table \ref{tab:meta-embeddings} and \ref{tab:meta-embeddings-centroids}.

\subsection{Word embedding representation of Tweets:}
For our standard machine learning models, we represent a Tweet by taking the average of the  vector of the individual words in that Tweet, ignoring the ones that are out of vocabulary. For our deep learning experiments, we take the vector of each word in a Tweet and concatenate them to get a word vector representation of the Tweet. Since this approach will not create a fixed length word vector representation, we pad each tweet to make their length equal to the maximum length Tweet in the training set. In the next sections we provide detailed technical descriptions of our word experimental setup. 


\section{Experimental setup} 

We experiment with all the
28 combinations from seven feature extraction methods, such as, BOW, NHEL, LIWC, TE, DE, ATE, ATE(orig.) and four standard machine learning models, such as, Multinomial Na\"{\i}ve Bayes (NB), Logistic Regression (LR), Linear Support Vector Machine (LSVM) and Support Vector Machine with radial basis kernel function (RSVM). In addition, 
we run experiments on all our four word embeddings and a randomly initialized embedding representations combined with our deep learning model (cbLSTM) to further analyse the efficacy of our proposed word embeddings in deep learning setting. We run all these experiments in our datasets (i.e. Dataset1 and Dataset2).

 \subsubsection{Train-test splits:} For a single experiment, we split all our data into a disjoint set of training (70\% of all the data) and testing (30\% of all the data) (see Table \ref{tab:splits}). 

\begin{table}[!ht]
\small
\centering
\begin{tabular}{|c|c|c|}
\hline
\textbf{Datasets} & \textbf{Train} & \textbf{Test} \\ \hline
Dataset1 & 355(178) & 152(76) \\ 
Dataset2 & 6127(613) & 2626(263)  \\ \hline
\end{tabular}
\caption{Number of Tweets in the train and test splits for the two datasets. The number of depressive Tweets is in parenthesis.
}
\label{tab:splits}
\end{table}

We use stratified sampling so that the original distribution of labels is retained in our splits. Furthermore, with the help of 10-fold cross validation in our training set, we learn the best parameter settings for all our model-feature extraction combinations, except for those that require no such parameter tuning. We then find the performance of the best model on our test set. 

We have run 30 such experiments on 30 random train-test splits. Finally, we report the performance of our model-feature extraction combinations based on the Precision, Recall, and F1 score averaged over the test sets of those 30 experiments. 

\subsubsection{Standard machine learning model specific settings:} 
For the SVMs and LR, we tune the parameter, $C \in \{ 2^{-9}, 2^{-7}, \dots, 
2^5 \}$ and 
additionally, $\gamma \in \{ 2^{-11}, 2^{-9},\dots, 2^2 \}$ for the RSVM (see scikit-learn SVM \footnote{\url{http://scikit-learn.org/stable/modules/svm.html}} and LR \footnote{\url{http://scikit-learn.org/stable/modules/svm.html}} docs for further description of these parameters). We use min-max feature scaling for all our features.

\subsubsection{Deep learning model specific settings:} 
\begin{figure}[!ht]
\centering
\includegraphics[width=0.15 \textwidth] {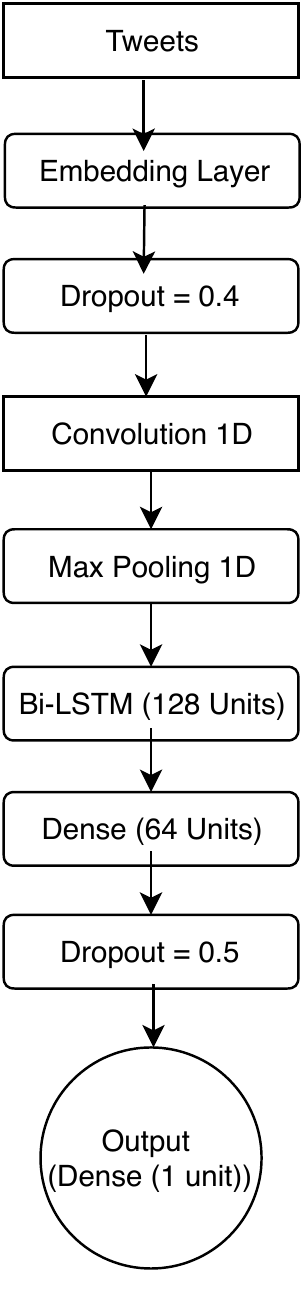} 
\caption{\label{fig:cbLSTM} Deep learning model (cbLSTM) architecture}
\end{figure}

We use a state of the art deep learning model which is a combination of Convolutional Neural Network (CNN) layer followed by a Bidirectional Long Short Term Memory (Bi-LSTM) layer (see Figure \ref{fig:cbLSTM}) inspired by the work of \cite{Zhou2015ACN} and \cite{nguyen2017deep}, which we name as cbLSTM.

From the train splits (as described in ``Train-test splits'' section), the deep learning model separates 10\% of samples for validation purpose and reports the result on test set. Although we have the liberty to learn our pre-trained word embedding in our deep learning model, we keep the embedding layer untrainable so that we can report the results that reflect only the effectiveness of pre-trained embedding, not the learned ones. Moreover, we report results on random initialized embedding to show how the other embeddings improved upon it. Since we have a smaller dataset, learning embedding do not introduce added value.

\section{Results analysis} \label{subsec:results}

\subsection{Quantitative performance analysis} 
Here we report the average results (i.e., average Precision, Recall and F1) for the best performing combinations among all the 28 combinations of our standard machine learning and feature extraction methods (as described in ``Experimental setup'' section). We also report the same for our four word embeddings combined with the deep learning (cbLSTM) model. We report these results separately for our Dataset1 and Dataset2.

Moreover, we report the results of two experiments, one by \cite{vioules2018detection} for Dataset1 and another by \cite{jamil2017monitoring} for Dataset2, where they use their own depression lexicons as a feature representation for their machine learning models. We report these two previous results because these are the most recent 
results on depressive Tweets identification task.  See Tables \ref{tab:dataset1-results} and \ref{tab:dataset2-results}.

\subsubsection{Standard machine learning models:}
\begin{figure}[!ht]
\centering
\includegraphics[width=0.50 \textwidth]{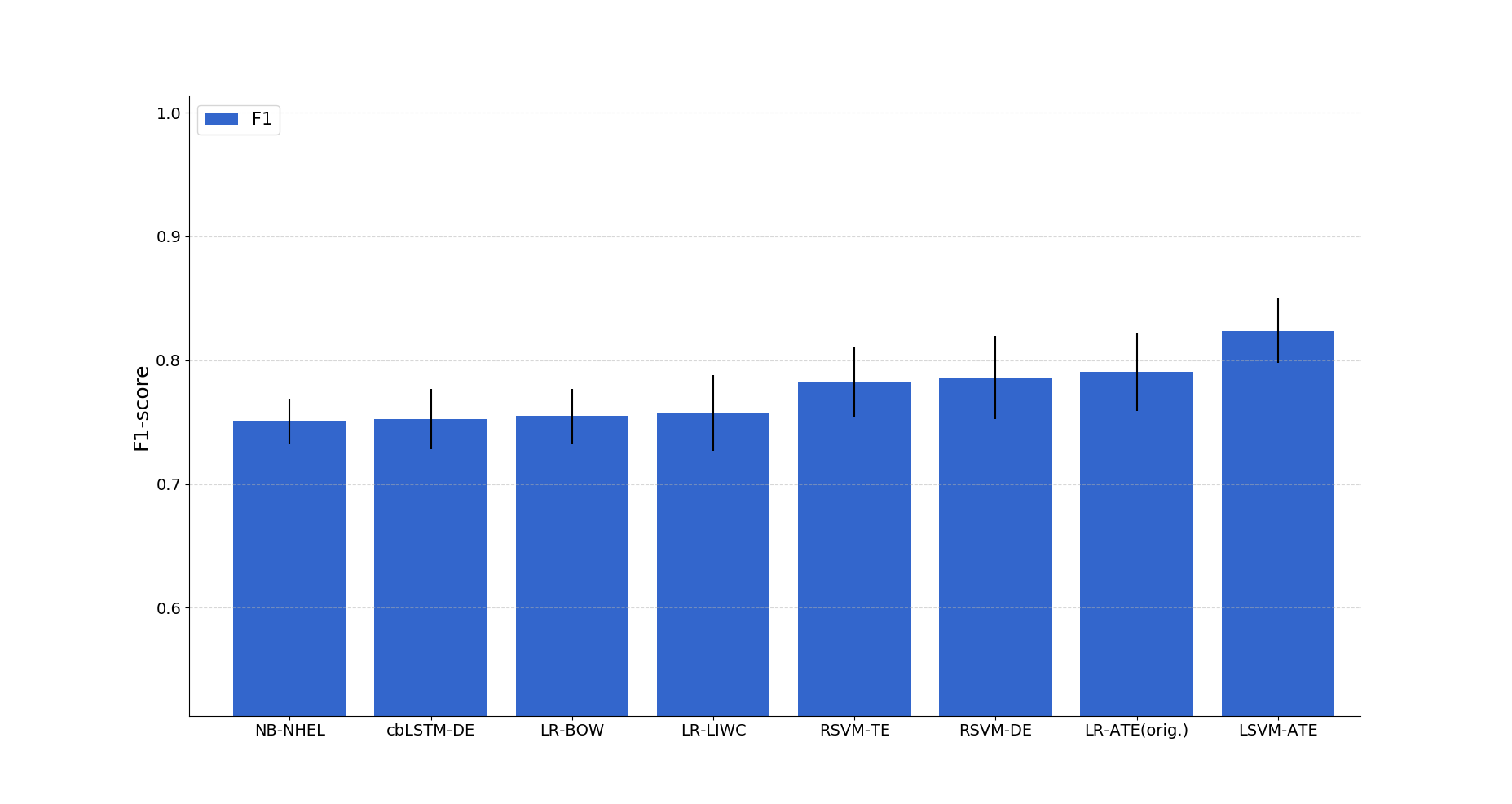}
\caption{\label{fig:barcharts-dataset1}  Error bars of F1 scores for our best model-feat. combinations on Dataset1}
\end{figure}

In general, Tweet level depression detection is a tough problem and a good F1 score is hard to achieve \cite{jamil2017monitoring}. Still, our LSVM-ATE achieves an average F1 score of $0.8238 \pm 0.0259$ which is 4\% better than our strongest baseline (RSVM-TE) with average F1 score of $0.7824 \pm 0.0276$ and 11\% better than \cite{vioules2018detection} with F1 score of 0.71 in Dataset1, see Table \ref{tab:dataset1-results} and Figure \ref{fig:barcharts-dataset1}.  

\begin{table}[!ht]
\small
\centering
\begin{tabular}{|c|C{2cm}|c|c|c|}
\hline
\textbf{Category} & \textbf{Model-Feat.} & \textbf{Prec.} & \textbf{Rec.} & \textbf{F1} \\ \hline
\multirow{1}{*}{Baselines} 
&NB-NHEL &0.6338	& \textbf{0.9224}	&0.7508   \\ 
&LR-BOW &0.6967  &0.8264  &0.7548 \\ 
 &LR-LIWC &0.7409	&0.7772	&0.7574  \\
 &RSVM-TE  &0.7739	&0.7939	&0.7824 \\
 \hline
\multirow{1}{*}{Our Models}
&cbLSTM-DE &0.6699	&0.8606	&0.7526 \\
&RSVM-DE &0.7495	&0.8280	&0.7859 \\ 
&LR-ATE(orig.) &0.7815	&0.8020	&0.7906 \\
&\textbf{LSVM-ATE} &\textbf{0.7984} &0.8520  &\textbf{0.8239}  \\ 
 
 \hline
 \multirow{1}{*}{Prev. Res.} &\cite{vioules2018detection} &0.71	&0.71	&0.71 \\ 
 \hline
\end{tabular}
 \caption{Average results on Dataset1 best model-feat combinations}
\label{tab:dataset1-results}
\end{table}

\begin{table}[!ht]
\small
\centering
\begin{tabular}{|c|C{2cm}|c|c|c|}
\hline
\textbf{Category} & \textbf{Model-Feat.} & \textbf{Prec.} & \textbf{Rec.} & \textbf{F1} \\ \hline
\multirow{1}{*}{Baselines} 
 &RSVM-NHEL &0.1754	&\textbf{0.7439}	&0.2858 \\ 
 &RSVM-BOW &0.2374	&0.5296	&0.3260  \\ 
 &RSVM-LIWC &0.2635	&0.6750	&0.3778 \\ 
 &RSVM-TE &0.3485	&0.6305	&0.4448 \\
 \hline
\multirow{1}{*}{Our Models} 
&RSVM-DE &0.3437	&0.5198	&0.4053 \\ 
&cbLSTM-ATE &\textbf{0.4416}	&0.3987	&0.4178 \\
&RSVM-ATE(orig.) &0.3476	&0.5648	&0.4276 \\
&\textbf{RSVM-ATE} &0.3675	&0.5923	&\textbf{0.4480}\\ 

 \hline
 \multirow{1}{*}{Prev. Res.} 
 &\cite{jamil2017monitoring} &0.1706 &0.5939 &0.265 \\ 
 \hline
\end{tabular}
 \caption{Average results on Dataset2 best model-feat combinations} 
\label{tab:dataset2-results}
\end{table}


\begin{table}[!ht]
\small
\centering
\begin{tabular}{|c|C{2cm}|c|c|c|}
\hline
\textbf{Category} & \textbf{Model-Feat.} & \textbf{Prec.} & \textbf{Rec.} & \textbf{F1} \\ \hline
\multirow{1}{*}{Baselines} 
 &cbLSTM-Random &0.5464	&\textbf{0.9817} &0.6986\\
 &cbLSTM-TE  &0.6510	 &0.8325	&0.7262 \\
 \hline
\multirow{1}{*}{Proposed} 
 &cbLSTM-ATE(orig.) &0.6288	&0.8439	&0.7093 \\
 &cbLSTM-ATE &\textbf{0.6915}    &0.8231	&0.7491 \\
 &\textbf{cbLSTM-DE} &0.6699	&0.8606	&\textbf{0.7526} \\
 \hline
\end{tabular}
 \caption{Average results for deep learning model (cbLSTM) in Dataset1 for all our (Baseline and Proposed) embeddings} 
\label{tab:dataset1-deep-learn}
\end{table}



\begin{table}[!ht]
\small
\centering
\begin{tabular}{|c|C{2cm}|c|c|c|}
\hline
\textbf{Category} & \textbf{Model-Feat.} & \textbf{Prec.} & \textbf{Rec.} & \textbf{F1} \\ \hline
\multirow{1}{*}{Baselines} 
&cbLSTM-Random &0.2308	&0.2791 &0.2502\\
&cbLSTM-TE &0.2615	&\textbf{0.6143}	&0.3655\\ 

 \hline
\multirow{1}{*}{Proposed} 
&cbLSTM-ATE(orig.) &\textbf{0.4598}	&0.3105	&0.3671 \\
&cbLSTM-DE &0.3231	&0.4891	&0.3880 \\
&\textbf{cbLSTM-ATE} &0.4416	&0.3987	&\textbf{0.4178} \\
 \hline
\end{tabular}
 \caption{Average results for deep learning model (cbLSTM) in Dataset2 for all our (Baseline and Proposed) embeddings} 
\label{tab:dataset2-deep-learn}
\end{table}

\begin{figure}[!ht]
\centering
\includegraphics[width=0.50 \textwidth]{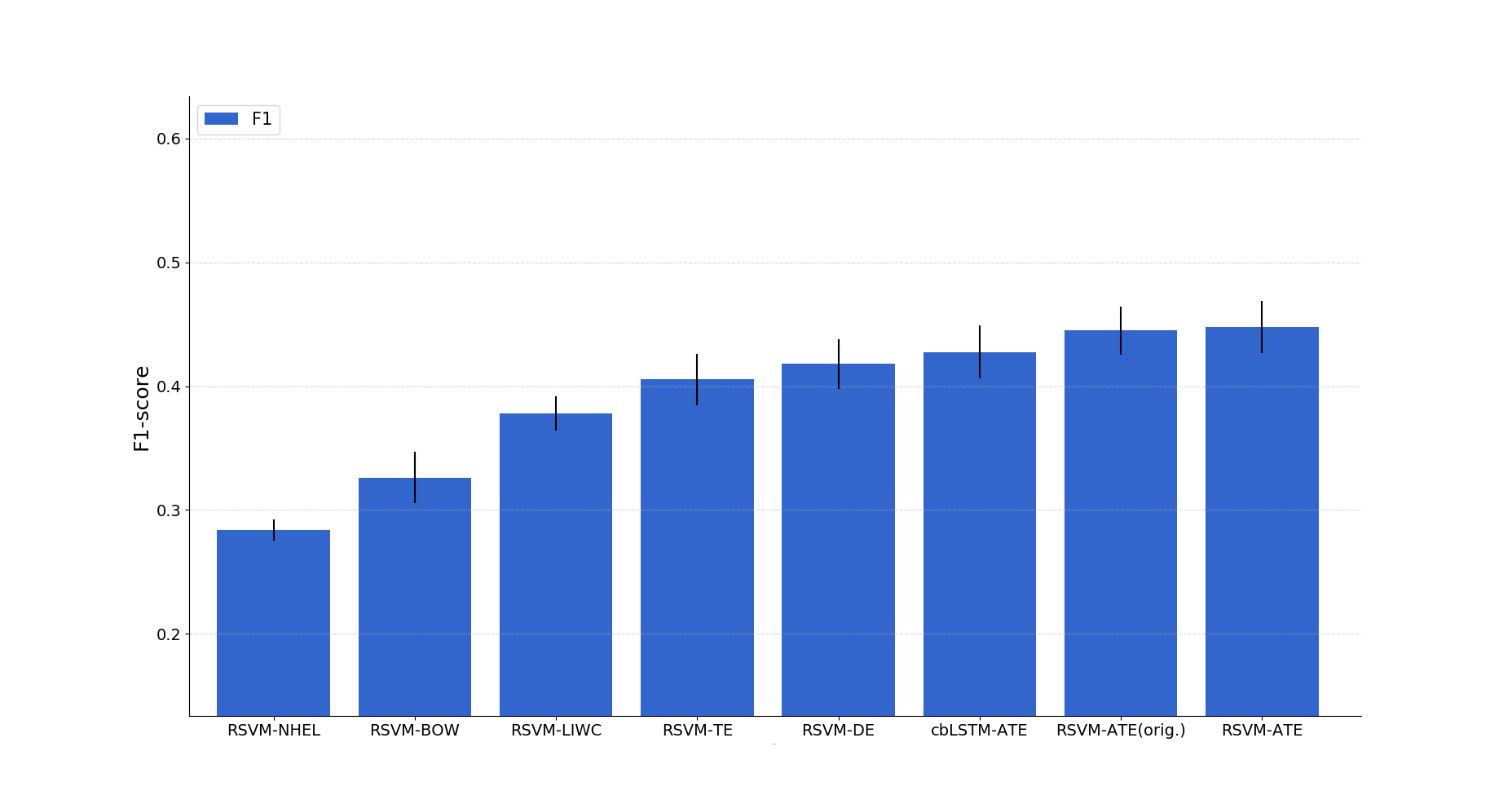}
\caption{\label{fig:barcharts-dataset2}  Error bars of F1 scores for our best model-feat. combinations on Dataset2}
\end{figure}

In Dataset2, which is imbalanced (90\% samples are non-depressive Tweets), our best model RSVM-ATE achieves 0.32\% better average F1 score (i.e. $0.4480 \pm 0.0209$)  than the strongest baseline, RSVM-TE with average F1 score of $0.4448 \pm 0.0197$  and 22.3\% better F1 score than \cite{jamil2017monitoring} (i.e. $0.265$) , see Table \ref{tab:dataset2-results} and Figure \ref{fig:barcharts-dataset2}.

In both datasets, NHEL has the best recall and the worst precision, while, BOW, LIWC and word embedding based methods have acceptable precision and recall.

\begin{figure}[!ht]
\centering
\includegraphics[width=0.45 \textwidth]{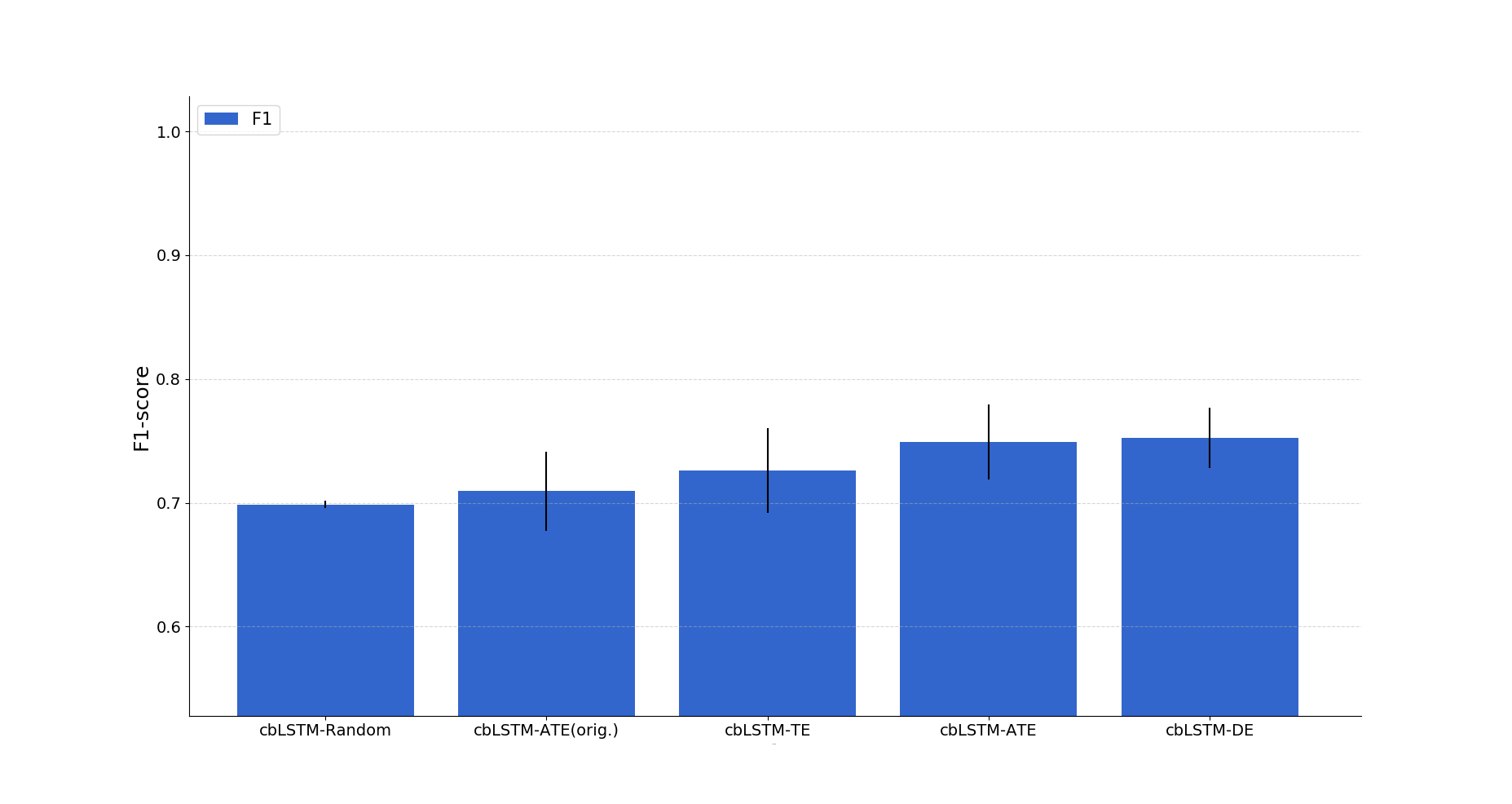}
\caption{\label{fig:barcharts-dataset1-cbLSTM}  Error bars of F1 scores for deep learning models (cbLSTM) with all our embeddings on Dataset1}
\end{figure}

\begin{figure}[!ht]
\centering
\includegraphics[width=0.45 \textwidth]{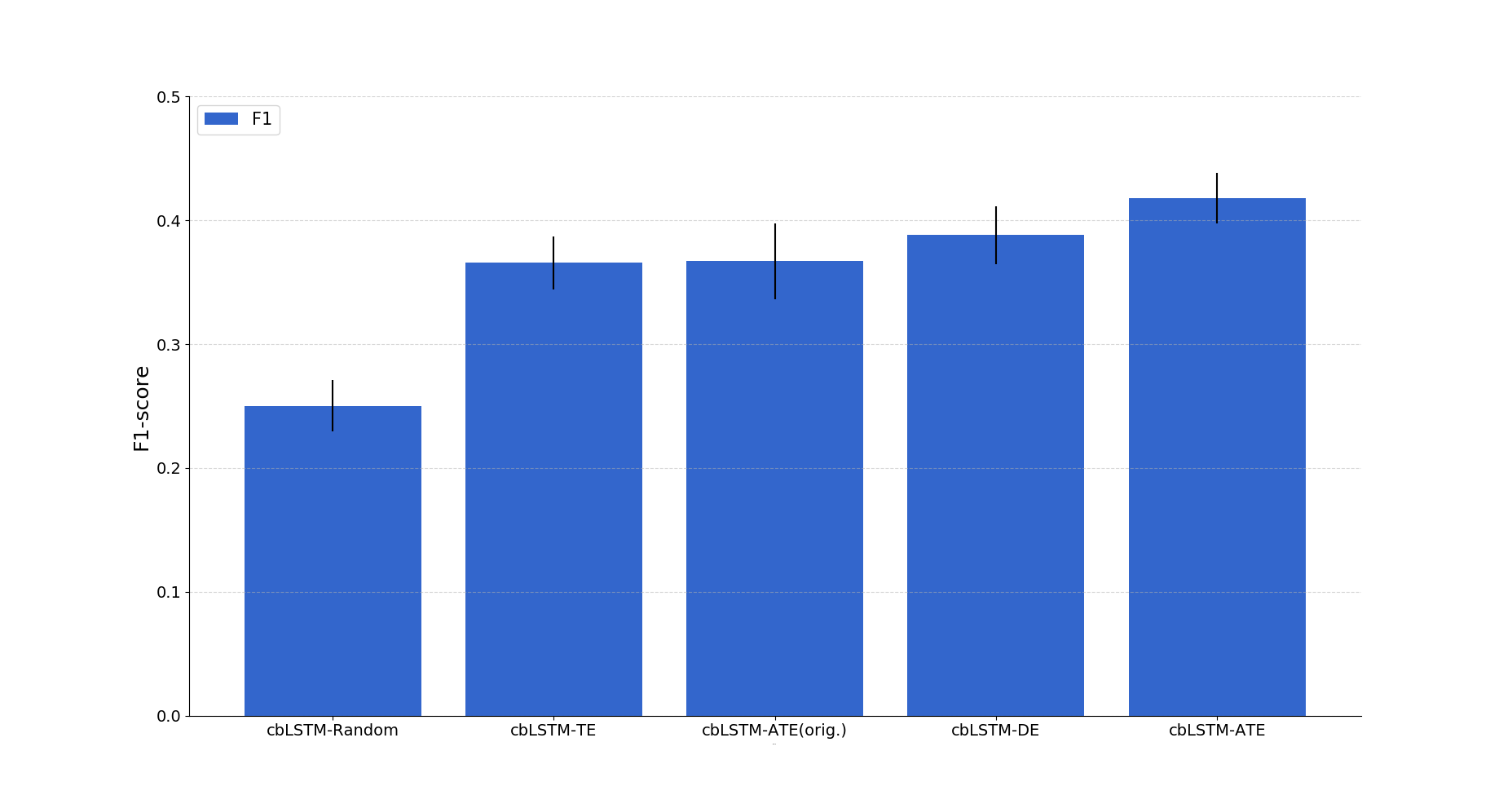}
\caption{\label{fig:barcharts-dataset2-cbLSTM}  Error bars of F1 scores for deep learning models (cbLSTM) with all our embeddings on Dataset2}
\end{figure}



\subsubsection{Deep learning model:}
In Dataset1, our best performing model cbLSTM-DE achieves around 3\% better in average F1 score (i.e. $0.7526 \pm 0.0244$) than the strongest baseline cbLSTM-TE with average F1 score of $0.7262 \pm 0.034$. It slightly performs better (0.35\%) than cbLSTM-ATE with average F1 score of $0.7491 \pm 0.0301$. In Dataset2, our best performing model cbLSTM-ATE with average F1 score of $0.4178 \pm 0.0202$  performs 5\% better than the strongest baseline cbLSTM-TE with average F1 score of $0.3655 \pm 0.021$ and around 2\% better than close contender cbLSTM-DE with average F1 score of $0.3880 \pm 0.0233$, see Figures \ref{fig:barcharts-dataset1-cbLSTM}, \ref{fig:barcharts-dataset2-cbLSTM} and Tables \ref{tab:dataset1-deep-learn}, \ref{tab:dataset2-deep-learn}. However, our best deep learning model performs on average 5\% lower in F1 scores than our standard machine learning models across our two datasets.
 
\subsection{Influence of datasets and feature representations in predictions}
\subsubsection{Standard machine learning models:}
Overall, in both datasets, word embedding based methods perform much better than BOW and lexicons. The reason is, 
they have a bigger vocabulary and better feature representation than BOW and lexicons. Among non-word embedding methods, BOW and LIWC perform better than NHEL, because the former provide better discriminating features than the latter. In Dataset1, ATE achieves better and stable F1 scores than both TE and DE with DE performing close enough. This confirms that DE can capture the semantics of depressive language very well. ATE is superior in performance because it leverages both the vocabulary coverage and semantics of a depressive language. In Dataset2, ATE achieves slightly better (although slightly unstable) F1 score than TE but significantly better F1 score than DE. The reason for this could be that the Tweet samples in Dataset2 are more about general distress than actual depression, also dataset is very imbalanced. In this case, the performance is affected mostly by the vocabulary size rather than the depressive language semantics. 
\subsubsection{Deep learning model:}
Although the overall F1 scores achieved with the deep learning model (cbLSTM) while used with the word embeddings is below par than the standard machine learning models for both datasets, we observe a general trend of improvement of our proposed ATE and DE embedding compared to strong baseline TE and random initialized embedding. Moreover this performance improvement is more pronounced in our noisy dataset (Dataset2), unlike standard machine learning models, suggests that deep learning models might be better at handling noisy data. Overall, all our proposed embeddings (i.e. DE, ATE) achieve more stable F1 scores than TE and random embeddings in both datasets. 

We believe the overall under-performance in the deep learning model is attributed to our small datasets. In our future analysis of the deep learning models and their performance we intend to experiment with a much larger and better dataset to derive more insights.

\subsection{Qualitative performance analysis} 
We report correctly predicted depressive Tweets in Table \ref{tab:correct_fn} by LSVM-ATE (our overall best model) 
which are mistakenly predicted as control Tweets (i.e., false negatives) when 
RSVM-TE (our strongest baseline) is used in a test set from Dataset1. The first example from Table \ref{tab:correct_fn}, \textit{``Tonight may definitely be the night''}, may be indicative of suicidal ideation and should not be taken lightly, also, the second one \textit{``0 days clean.''} is the trade mark indication of continued self-harm, although many depression detection models will predict these as normal Tweets. It is also interesting to see how our best word embedding is helpful in identifying depressive Tweets which are more subtle like, \textit{``Everyone is better off without me. Everyone.''}.

\begin{table}[!ht]
\small
\centering
\begin{tabular}{|p{7.4cm}|}
\hline
\textbf{Tweets} \\ \hline
``Tonight may definitely be the night.'' \\ \hline
``0 days clean.'' \\ \hline
``Everyone is better off without me. Everyone.'' \\ \hline
``Is it so much to ask to have a friend who will be there for you no matter what?'' \\ \hline
``I understand you're `busy', but fuck that ... people make time for what they want.'' \\ \hline
`` I'm a failure.'' \\ \hline
\end{tabular}
\caption{False negative depressive Tweets when TE is used, correctly predicted when ATE is used in a test set from Dataset1.}
\label{tab:correct_fn}
\end{table}

\begin{figure}[!ht]
\centering
\includegraphics[width=8.59cm] {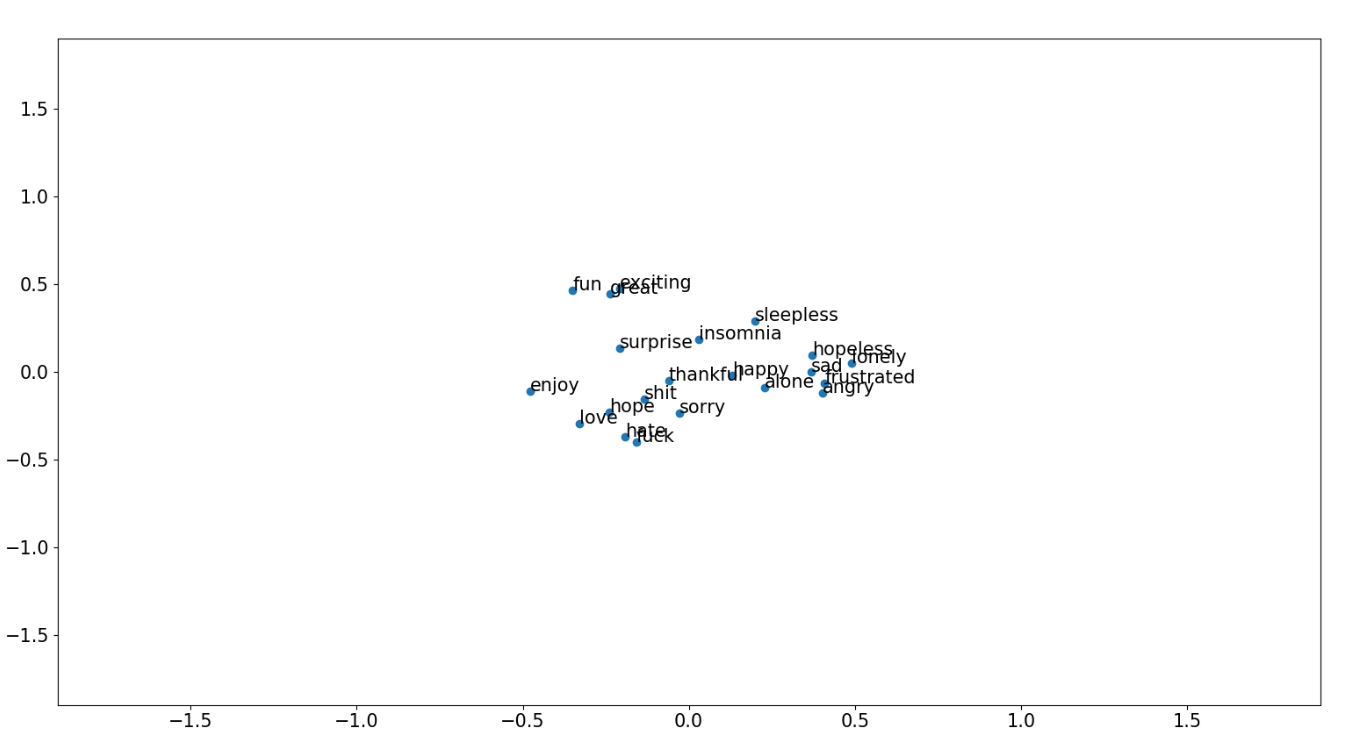} 
\caption{\label{fig:GTEcluster} Two-dimensional PCA projection of LIWC 
POSEMO and NEGEMO words (frequently occured in our datasets) in General Twitter word Embedding (TE).}
\end{figure}

\begin{figure}[!ht]
\centering
\includegraphics[width=8.59cm] {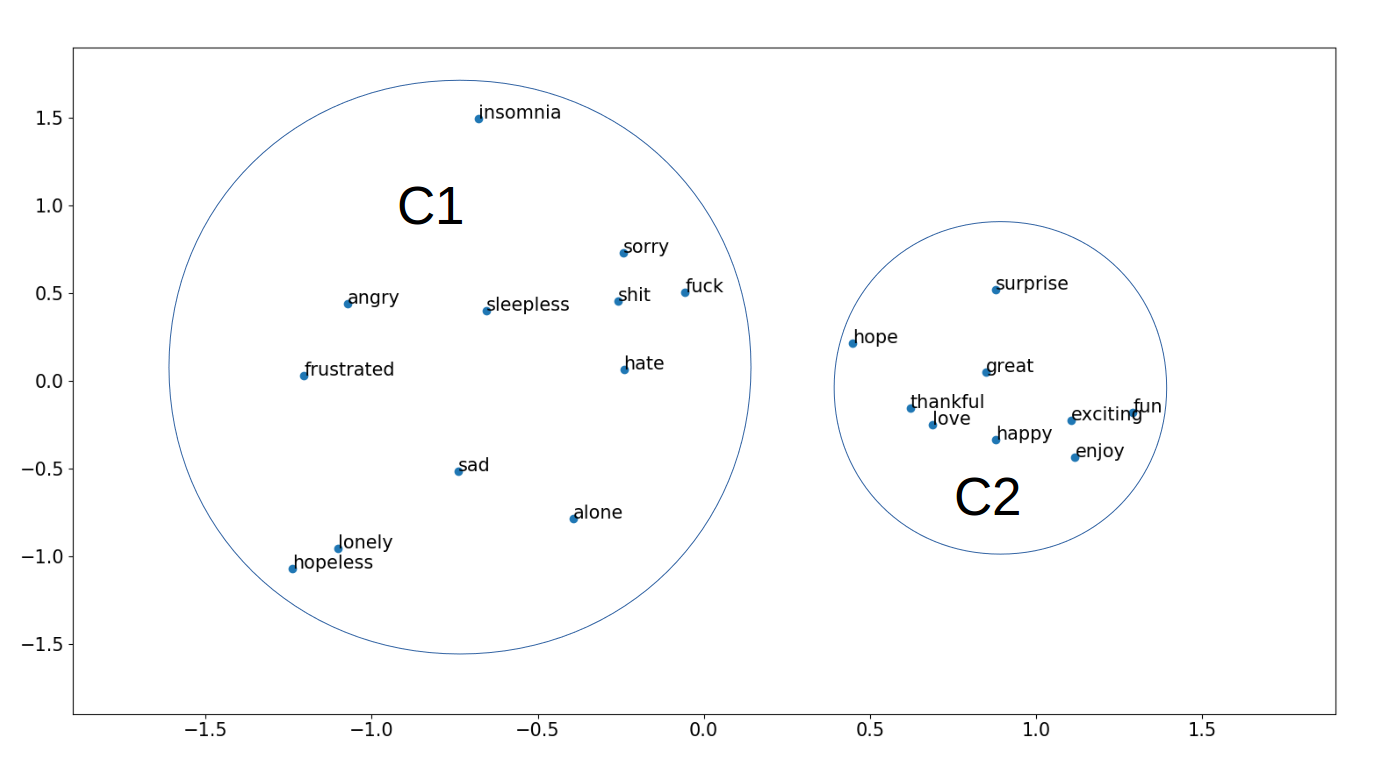} 
\caption{\label{fig:ATEcluster} Two-dimensional PCA projection of LIWC POSEMO and NEGEMO words (frequently occured in our datasets) in Adjusted Twitter word Embedding (ATE).}
\end{figure}

According to earlier research, depression has close connection with abnormal regulation of positive and negative emotion \cite{kuppens2012emotional} and \cite{seabrook2018predicting}. So to consider how the words carrying positive and negative sentiment are situated in our adjusted vector space, we plot the PCA projections of ATE and TE for 
the high frequency words used in both datasets that are the members of LIWC positive emotion (POSEMO) and negative emotion
(NEGEMO) 
categories.  We observe that POSEMO and NEGEMO words form two clearly distinctive clusters, i.e., C2 and C1 respectively in ATE. We also notice the word ``insomnia'' and ``sleepless'' which represent common sleep problem in depressed people, reside in C1 or NEGEMO cluster. However, we do not see any such clusters in TE. See Figure \ref{fig:ATEcluster} and \ref{fig:GTEcluster}.  We believe this distinctions of affective contents in vector space partially play a role in our overall accuracy. Also the PCA projection gives a glimpse of the semantic relationship of affective words in depressive language. Although its not an exhaustive analysis but a insightful one that we believe would be helpful for further analysis of affect in depressive language.

\subsection{Effects of embedding augmentation:}
The fact that generally ATE performs better (see Tables \ref{tab:dataset1-results}, \ref{tab:dataset2-results}, \ref{tab:dataset1-deep-learn} and \ref{tab:dataset2-deep-learn}) than TE proves that our embedding adjustment improves the semantic representation of TE, because ATE and TE both have exactly the same vocabulary. Further to show that embedding adjustment for OOVs contributes to this improved accuracy, we run an experiment, where we replace the words in TE which are common to DE, with their ATE word vectors. We name this embedding, ``$TE \cap DE$-in-TE-adjusted.'' This confirms that none of the OOVs are adjusted and we compare this result with our ATE (where all the OOVs are adjusted). In this experiment, we see $TE \cap DE$-in-TE-adjusted obtains 2.62\% less F1 score compared to ATE for Dataset1 and 2.13\% less F1 score for the same in Dataset2, confirming semantic adjustment of OOVs does play an important role. Also, this model has less stable F1 scores than our best model in both datasets. See Table \ref{tab:embed-augment}


\begin{table}[!ht]
\small
\centering
\begin{tabular}{|c|C{3.5cm}|c|}
\hline
\textbf{Dataset} & \textbf{Model-Feat.} &\textbf{F1} \\ \hline
\multirow{1}{*}{Dataset1} 
&LR-$TE \cap DE$-in-TE-adjusted  & $0.7977 \pm 0.0305$ \\ 
&\textbf{LSVM-ATE (Best Model)} &$\bm{0.8239 \pm 0.0259}$ \\
 \hline
\multirow{1}{*}{Dataset2} 
 &RSVM-$TE \cap DE$-in-TE-adjusted  &$0.4267 \pm 0.0217$ \\ 
 &\textbf{RSVM-ATE (Best Model)} &$\bm{0.4480 \pm 0.0209}$ \\
 \hline
\end{tabular}
 \caption{Average F1 scores from experiments to show the effect of our augmentation for OOVs}
\label{tab:embed-augment}
\end{table}


\section{Ethical concerns} \label{subsec:ethics} We use \textit{Suicidal Forum} posts where users are strictly required to stay anonymous. Additionally, it employs active moderators who regularly anonymize the contents in case the user reveals something that can identify them. Moreover, we use Reddit and Twitter public posts which incur minimal risk of user privacy violation as established by earlier research (\cite{milne2016clpsych}, \cite{coppersmith2015adhd} and \cite{losada2016test}) utilizing the same kind of data. We also obtained our university ethics board approval that lets us use datasets collected by external organizations.

\section{Conclusion} 
In this paper, we empirically present the following observations for a high quality dataset:

\begin{itemize}
\item  For depressive Tweets detection, we can use word embedding trained in an unsupervised manner on a small corpus of depression forum posts, which we call Depression specific word Embedding (DE), and then use it as a feature representation for our machine learning models. This approach can achieve good accuracy, despite the fact that it has 100 times smaller vocabulary than our general Twitter pre-trained word embedding.
\item Furthermore, we can use DE to adjust the general Twitter pre-trained word Embedding (available off the shelf) or TE through non-linear mapping between them. This adjusted Twitter word Embedding (ATE) helps us achieve even better results for our task. 
\item We need not to depend on human annotated data or labelled data for any of our word embedding representation creation.
\item Depression forum posts have specific distributed representation of words and it is different than that of general twitter posts and this is reflected in ATE, see Figure \ref{fig:ATEcluster}. 

\item We intend to make our depression corpus and embeddings publicly available upon acceptance of our paper. 

\end{itemize}

\section{Future work}
 In the future, we would like to analyze DE more exhaustively to find any patterns in semantic clusters that specifically identify depressive language. We would also like to use ATE for Twitter depression lexicon induction and for discovering depressive Tweets. We can see great promise in its use in creating a semi-supervised learning based automated depression data annotation task later on.

\section{Supplementary Results}
Here we report Tables \ref{tab:other-vocab}, \ref{tab:ate-centroids}, \ref{tab:meta-embeddings} and \ref{tab:meta-embeddings-centroids} for the analysis of our proposed augmentation methods.

\begin{table}[!htp]
\small
\centering
\begin{tabular}{|c|c|C{2.2cm}|}
\hline
\textbf{Dataset} &\textbf{Model-Feat.} &\textbf{F1} \\ \hline
\multirow{1}{*}{Dataset1}
&\textbf{LSVM-ATE (Best Model)} &\textbf{$\bm{0.8239 \pm 0.0259}$} \\ 
&RSVM-DE &$0.7859 \pm 0.0335$ \\
&RSVM-TE &$0.7824 \pm 0.0276$ \\

&RSVM-$ATE \cap DE$ &$0.7786 \pm 0.0290$ \\
&LSVM-$TE \cap DE$ &$0.7593 \pm 0.0340$\\
&RSVM-$TE\cap DE$-concat &$0.7834 \pm 0.0321$ \\ 
&LR-$TE \cap DE$-in-TE-adjusted  &$0.7977 \pm 0.0304$ \\ 
\hline
\multirow{1}{*}{Dataset2} 
&\textbf{RSVM-ATE (Best Model)} &\textbf{$\bm{0.4480 \pm 0.0209}$} \\ 
&RSVM-DE &$0.4351 \pm 0.0206$ \\
&RSVM-TE &$0.4448 \pm 0.0196$ \\
&RSVM-$ATE \cap DE$ &$0.4155 \pm 0.028$ \\
&LSVM-$TE \cap DE$ &$0.4002 \pm 0.0292$\\
&RSVM-$TE\cap DE$-concat &$0.4249 \pm 0.0187$ \\ 
&RSVM-$TE \cap DE$-in-TE-adjusted  &$0.4267 \pm 0.0216$ \\ 
\hline
\end{tabular}
\caption{Average Precision, Recall and F1 scores from experiments to show the effect of our augmentation}
\label{tab:other-vocab}
\end{table}

\begin{table}[!htp]
\small
\centering
\begin{tabular}{|c|c|c|}
\hline
\textbf{Dataset} &\textbf{Model-Feat.} &\textbf{F1} \\ \hline
\multirow{1}{*}{Dataset1} 
&\textbf{LSVM-ATE (Best Model)} &\textbf{$\bm{0.8239 \pm 0.0259}$} \\ 
&LR-ATE-Centroid-1  &$0.8219 \pm 0.0257$ \\
&LR-ATE-Centroid-2  &$0.8116 \pm 0.029$\\
\hline
\multirow{1}{*}{Dataset2} 
&\textbf{RSVM-ATE (Best Model)} &\textbf{$\bm{0.4480 \pm 0.0209}$} \\ 
&RSVM-ATE-Centroid-1   &$0.4372 \pm 0.0172$  \\
&RSVM-ATE-Centroid-2  &$0.4431 \pm 0.0211$ \\
\hline
\end{tabular}
\caption{\label{tab:ate-centroids}Experiments on centroid based methods}

\end{table}

\begin{table}[!htp]
\small
\centering
\begin{tabular}{|c|c|c|}
\hline
\textbf{Dataset} &\textbf{Model-Feat.} &\textbf{F1} \\ \hline
\multirow{1}{*}{Dataset1} 
&\textbf{LSVM-ATE (Best Model)} &\textbf{$\bm{0.8239 \pm 0.0259}$} \\ 
&NB-ATE-AAEME &$0.7845 \pm 0.0229$   \\ 
&LR-ATE-AAEME-OOV &$0.8014 \pm 0.0333$  \\
\hline
\multirow{1}{*}{Dataset2}
&\textbf{RSVM-ATE (Best Model)} &\textbf{$\bm{0.4480 \pm 0.0209}$} \\ 
&RSVM-ATE-AAEME &$0.4166 \pm 0.0187$   \\ 
&RSVM-ATE-AAEME-OOV &$0.4433 \pm 0.0200$ \\
\hline
\end{tabular}
\caption{\label{tab:meta-embeddings}Experiments on meta-embedding methods}

\end{table}

\begin{table}[!htp]
\small
\centering
\begin{tabular}{|c|c|C{2.2cm}|}
\hline
\textbf{Dataset} &\textbf{Model-Feat.} &\textbf{F1} \\ \hline
\multirow{1}{*}{Auto-Encoder Based Methods} 
&\textbf{ATE (Best Model)} &\textbf{$\bm{0.8239 \pm 0.0259}$} \\ 
&ATE-AAEME &$0.7845 \pm 0.0229$   \\ 
&ATE-AAEME-OOV &$0.8014 \pm 0.0333$  \\
\hline
\multirow{1}{*}{Centroid Based Methods}
&ATE-Centroid-1  &$0.8219 \pm 0.0257$ \\
&ATE-Centroid-2  &$0.8116 \pm 0.029$\\ 
\hline
\end{tabular}
\caption{\label{tab:meta-embeddings-centroids}Experiments on meta-embedding and centroid based methods on Dataset-1}
\end{table}
 
\bibliography{emnlp2018}
\bibliographystyle{aaai}
\end{document}